\documentclass[twoside,11pt,table]{article}

\usepackage{graphicx}
\usepackage{comment}
\usepackage{booktabs}
\usepackage{dirtytalk}
\usepackage{xcolor}
\usepackage[bookmarks=true]{hyperref}
\hypersetup{%
    bookmarks=false,    
    pdftitle={TeX and friends hackers manual},    
    pdfauthor={Yiannis Lazarides},                     
    pdfsubject={TeX and LaTeX},                        
    pdfkeywords={TeX, LaTeX, graphics, images}, 
    colorlinks=true,       
    linkcolor=blue,       
    citecolor=black,       
    filecolor=black,        
    urlcolor=purple,        
    linktoc=page            
}
\usepackage{cleveref}

\usepackage{amssymb}
\usepackage{wasysym}
\usepackage{multirow}
\usepackage{makecell}
\usepackage[inline,shortlabels]{enumitem}

\usepackage{array}
\usepackage{longtable}

\usepackage{jair, theapa, rawfonts}
\newcommand\citep{\cite}	
\newcommand\citet{\citeA}	
\renewcommand{\cite}{\shortcite}
\renewcommand{\citet}{\shortciteA}

\makeatletter
\newcommand{\noindentparagraph}{%
  \@startsection{paragraph}{4}%
  {\z@}{1.0ex \@plus 1ex \@minus .2ex}{-1em}%
  {\normalfont\normalsize\bfseries}%
}

\definecolor{lightgray}{gray}{0.95}
\let\oldtabular\tabular
\let\endoldtabular\endtabular
\renewenvironment{tabular}{\rowcolors{2}{white}{lightgray}\oldtabular}{\endoldtabular}
\newcommand{\ra}[1]{\renewcommand{\arraystretch}{#1}}

\jairheading{73}{2023}{459-489}{07/2022}{11/2023}
\ShortHeadings{Diagnosing AI Explanation Methods with Folk Concepts of Behavior}
{Jacovi, Bastings, Gehrmann, Goldberg \& Filippova}
\firstpageno{459}

\begin{document}

\title{Diagnosing AI Explanation Methods with \\ Folk Concepts of Behavior}

\author{\name Alon Jacovi \email alonjacovi@gmail.com \\
       \addr Bar Ilan University
       \AND
       \name Jasmijn Bastings \email bastings@google.com \\
       \addr Google DeepMind
       \AND
       \name Sebastian Gehrmann \email sgehrmann8@bloomberg.net \\
       \addr Google Research
       \AND
       \name Yoav Goldberg \email yoav.goldberg@gmail.com \\
       \addr Bar Ilan University, \\ Allen Institute for Artificial Intelligence 
       \AND
       \name Katja Filippova \email katjaf@google.com \\
       \addr Google DeepMind}


\maketitle

\begin{abstract}
We investigate a formalism for the conditions of a successful explanation of AI. We consider ``success'' to depend not only on what information the explanation contains, but also on what information the human explainee understands from it.
Theory of mind literature discusses the folk concepts that humans use to understand and generalize behavior. We posit that folk concepts of behavior provide us with a ``language'' that humans understand behavior with. We use these folk concepts as a framework of \textit{social attribution} by the human explainee---the information constructs that humans are likely to comprehend from explanations---by introducing a blueprint for an explanatory narrative (\Cref{fig:teaser}) that explains AI behavior with these constructs. 
We then demonstrate that many XAI methods today can be mapped to folk concepts of behavior in a qualitative evaluation. This allows us to uncover their failure modes that prevent current methods from explaining successfully---i.e., the information constructs that are missing for any given XAI method, and whose inclusion can decrease the likelihood of misunderstanding AI behavior.
\end{abstract}

\begin{figure}[t]
\centering
  \includegraphics[width=0.95\textwidth]{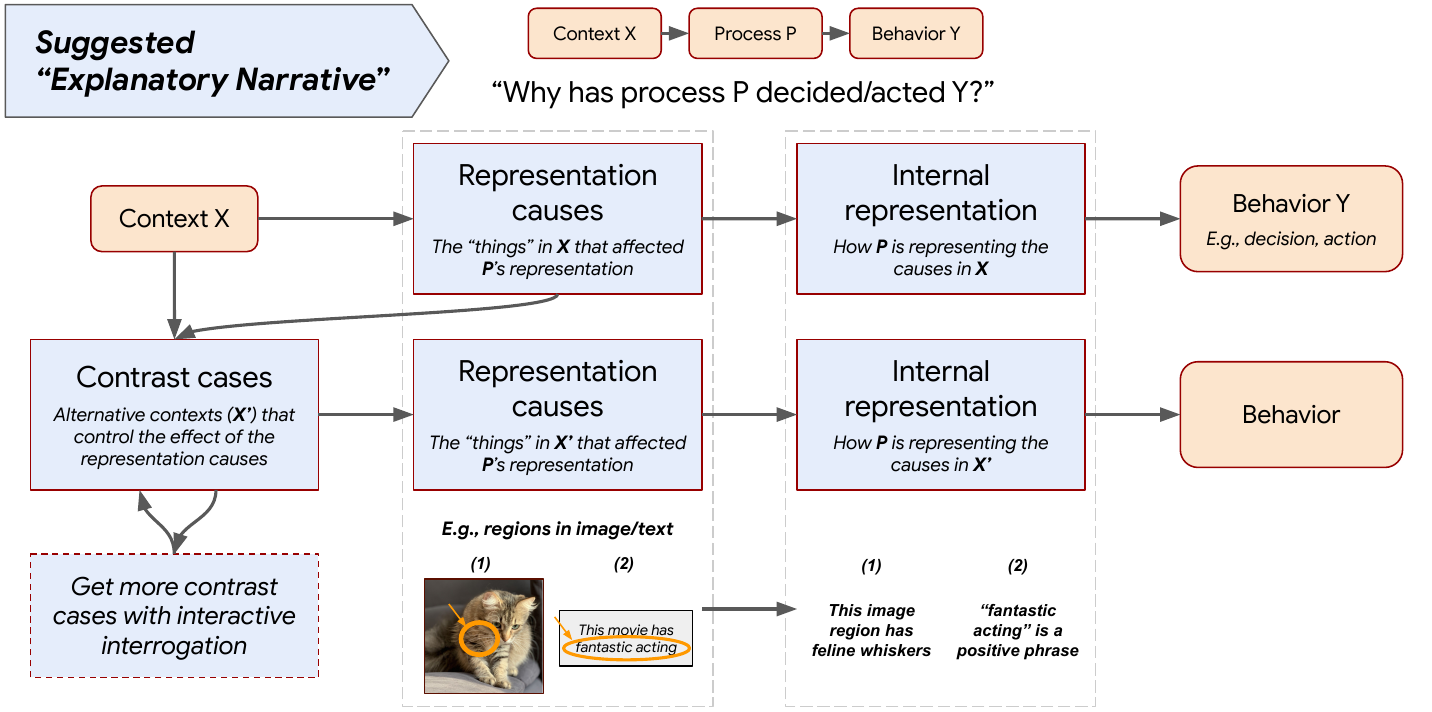}
  \caption{Schematic of an explanatory narrative, as explored in this work. The narrative communicates a causal chain composed of two categories of causes: The objective causes in the context, and the actors' subjective interpretation of those causes. The causes' role is communicated against alternative contexts (contrast cases) that intervene on them. 
  This paper develops the narrative structure, justifies it with precedence, and applies it to modern XAI methodologies to derive useful insights about the path towards successfully explaining complex AI.
  }
  \label{fig:teaser}
\end{figure}

\section{Introduction}
\label{Introduction}

In the development of explanation methods of AI systems, there is often a strong focus on the side of the explainer, but little attention is being paid to the exchange of information between the explainer and the explainee~\cite{diogo2019-survey,DBLP:journals/ki/SokolF20,DBLP:journals/corr/abs-2101-09429}. In particular, when explanation methods are introduced, they are typically motivated by being able to satisfy certain mathematical properties, which are not necessarily grounded in the needs of the explainee~\cite{miller19-social,rutjes2019-chi}. Yet, explainees have different experiences and expertise and may thus not understand an explanation in the intended way.
We aim to formalize what explainees may understand about AI processes as a result of explanations, and how this understanding may differ from what the explanation attempted to communicate. We refer to the information which the explainee comprehends as the explainee's mental model. In this work, we are concerned with AI explanations as \textit{explanations describing AI processes} (``why did this behavior manifest?''), rather than explanations as justifications to AI decisions (``why is this behavior correct?'')  \cite{DBLP:journals/corr/abs-2010-12896,make3040045}.

XAI methods can fulfill one or more desiderata for what explanations ``should'' satisfy: For example, that explanations should accurately describe the AI system they are explaining~\cite{DBLP:conf/dsaa/GilpinBYBSK18,DBLP:journals/corr/abs-1811-10154,DBLP:conf/aies/LakkarajuKCL19}; or be sufficient, so that no crucial information is missing~\cite{DBLP:conf/emnlp/YuCZJ19,DBLP:journals/entropy/LinardatosPK21}; or be minimal, so that no redundant information is given~\cite{DBLP:conf/emnlp/LeiBJ16,DBLP:journals/entropy/LinardatosPK21}; and so on~\cite{DBLP:journals/cacm/Lipton18}. Such constraints are given mathematical form, and then argued for by demonstrating that XAI methods which do \textit{not} uphold the mathematical constraints, fail in some core utility~\cite{melis2018-robustness-of-interpretability,feng-etal-2018-pathologies,DBLP:journals/corr/abs-1907-00570}.

This method of selecting desiderata is appealing in its flexibility, but without knowing the cognitive principles behind such desiderata, 
which are often born from AI practitioners' intuitions, we cannot say for certain whether, or \textit{why}, they are necessary or useful properties of explanations that successfully communicate information.\footnote{While such questions can be answered through user studies and a better understanding of user experiences and mental models, researchers often put 
explanations in the hands of unknown users through the release of tools~\cite<e.g.,>{tenney2020language,kokhlikyan2020captum} without knowing how their users will interpret the results, regardless of the axioms that are being satisfied. 
Moreover, even when user studies are conducted, studying explanations in isolation is not a replacement for studying them after their deployment in actual systems \cite{DBLP:conf/iui/BucincaLGG20}.} 

In \cref{sec:faithfulness} we address this problem by characterizing ``successful explanations'' in a consistent framework, rather than via a set of axioms (e.g., faithfulness or sufficiency), by pivoting the root of the analysis from the formal properties of the explainer to the cognitive properties of the explainee. The set of desiderata of a successful explanation is then \textit{derived from} this foundational framework, as desiderata that enable a specific formalism of success---coherence---rather than being treated as axioms. 
To develop our framework, we draw inspiration from psychological and philosophical research in the field of \textit{theory-theory}~\cite{Morton1980-MORFOM-2}: The study of how humans model the outside world for the purpose of generalizing, understanding and explaining phenomena. 

Research in this area additionally points to \textit{biases}, or \textit{habits}, that a human explainee commonly exhibits when leveraging prior knowledge in comprehending explanations of behavior. One of these habits is to understand non-human processes by drawing analogies to human behavior (\Cref{sec:anthropomorphism-intent}). This makes the area of \textit{theory of mind} and \textit{folk psychology}, the study of how humans model human behavior, relevant to characterize humans' understanding of AI explanations and AI behavior \cite{miller19-social}. In \Cref{subsec:folk-concepts} we investigate how folk concepts of behavior can help us to construct causal explanations of behavior.

We apply this framework to the XAI literature (\Cref{sec:analysis-xai-methods}), and find that many XAI mechanisms can be aligned with these folk concepts of behavior. This lets us identify what different XAI methods are \textit{missing} which could potentially increase their ability to communicate information successfully. 
We do this by observing what the explanation method fails to communicate among the causal explanatory narrative in \Cref{fig:teaser}. The missing component is in danger of being incorrectly speculated upon by the explainee, which can cause the explainee to misunderstand the explanation.

Below we summarize our primary findings:
\begin{enumerate}
    \item We say that behavior has been successfully explained if the explainee's mental model is \textit{coherent}, in that no contradictions are found between it and additional observations of behavior (\Cref{sec:faithfulness}). 
    \item An explainee's understanding of behavior can be conceptualized with multiple components: The internal representation of the behaving actor(s), the things that affected this representation, and the things that affected the outcome without affecting the representation. If any components are missing, the explainee may incorrectly assume them, leading to a misunderstanding of the explained process (\Cref{sec:folk-psychology-model}).
    \item We show that for a wide variety of current explanation methods, each of them fails the completeness test---i.e., is missing at least one of the components that compose the explanatory narrative (\Cref{sec:analysis-xai-methods}).
\end{enumerate}

Based on these findings, we conclude that to minimize the possibility of misunderstanding by human explainees' in XAI, there are two tools at our disposal (\Cref{sec:diagnosis-conclusion}): (1) Communicating information via a ``complete'' causal narrative; (2) Using interactivity to explore contrast cases that intervene on the narrative's suggested causes, as a medium of resolving contradictions methodologically.

\section{Defining Successful Explanations} \label{sec:faithfulness}



When comprehending an explanation to some event, the human explainee establishes a hypothesis about the event's history. We refer to this as the explainee's \textit{mental model} \cite{payne2003-mental-model}. Viewing the explanation as a ``function'' \cite{Lombrozo2006-LOMTSA}, the mental model can be considered as the outcome of the explanation. The conditions to a successful explanation are therefore conditions about the explainee's mental model \cite{SREEDHARAN2021103558}.

\subsection{Coherent Mental Models}

The cognitive science literature\footnote{We focus on the cognitive and developmental function of explanation in humans, as opposed to the social utility of explanation, as the uses of explanation in society (e.g., teaching, assigning blame) build on this core cognitive function.} often describes the goal of an explanation for the explainee as \textit{generalization and prediction}~\cite{10.2307/3506202,Lombrozo2006-LOMTSA,williams2010-explanation-role,bradley2017-omit}. This means that an explainee develops a ``coherent'' hypothesis about the circumstances that led to the explained event which is consistent even for new events~\cite{murphy1985,johnson2002conditionals}, and enables them to make predictions about these events (also known as \textit{explanatory unification,} \cite{Kitcher1981-KITEU} and \textit{consilience} \cite{10.7551/mitpress/1968.001.0001}). In the case of AI, this means generalizing to other instances of AI behavior. Therefore, in this work we consider an explanation as ``successful'' if it produces a mental model which is coherent across instances of AI behavior.

The principal constraint posed by coherency is that there are no contradictions between the explainee's mental model and additional observations (contrast cases).
For example, when hiding a ball under a cup, the theory that the ball continues to exist is consistent with (does not contradict) the new observation of the ball when removing the cup. This insight relies on the explainee's mental model of object permanence.\footnote{In the more complex context of AI, a similar understanding can be facilitated through training programs and instructional aid that shape mental models of humans about AI behavior~\cite{hanisch1991cognitive,DBLP:journals/tvcg/GehrmannSKPR20}.}

\subsection{Implications}

The definition of successful explanation as a function of coherent mental models implies several interesting conclusions:

\noindentparagraph{Explanation ``correctness'' is not explicitly part of this definition.} 
A recent trend of the XAI evaluation literature pertains to the \emph{correctness of the explanation with respect to the AI}: Whether an explanation faithfully represents information about the model. The literature in this area establishes that XAI methods, as lossy approximations of the AI's reasoning process, are not completely faithful~\cite{DBLP:journals/corr/abs-1810-03292,ghorbani2019-fragile-interpretability} and that completely faithful and human-readable explanations are likely an unreasonable goal~\cite{jacovi2020-faithfully}. Various relaxed measures of faithfulness were proposed (Appendix \ref{app:faithfulness-coherence-alignment}). 

However, human-to-human explanations also often do not provide correctness guarantees and yet are common and accepted.
While an explanation should not ``incorrectly'' describe the event history, some allowance is permitted on the uncertainty of whether the explanation is considered correct, in the absence of ground truth. This allowance manifests by using \textit{coherence}, rather than \textit{correctness}, due to this intractability. In Appendix \ref{app:faithfulness-coherence-alignment} we discuss how empirical quality measures of AI explanations, developed in recent years by the XAI community, also in fact capture coherence.

In this perspective, correctness or faithfulness can be considered to be a useful \textit{but not necessary} condition to successful explanations---since faithfulness can contribute to coherence, but is not the only means of doing so. 

\noindentparagraph{Coherence is characterized by an \textit{empirical} budget allotted to proving or refuting it.} Coherence positions the success metric of an explanation as an empirical measure rather than a theoretical one. If no contradiction is found after a ``sufficient enough'' search, an explanation is deemed ``correct enough'' \cite{Sellars1963-SELSPA-2,Kitcher1981-KITEU,Lehrer1990-LEHTOK,theoriesofexplanation-mayes}.\footnote{Precise affordances of this budget is beyond our scope, and should be considered societal or regulatory in nature. For use cases with large state spaces, e.g., language generation or reinforcement learning, the problem of summarizing agent behavior under a constrained budget has been studied~\cite<e.g.,>{DBLP:journals/aamas/AmirDS19}.} 

\noindentparagraph{Explanation is interactive: Lack of coherence---the existence of contradictions---is \textit{not} a failure state.} The explainee establishes a mental model as a result of explanation via an \textit{iterative} process, rather than one-time. This means that if coherence was refuted, i.e. contradictions arise, the mental model is deemed insufficient and can be \textit{adjusted} by the explainee into one for which the contradiction is resolved \cite{Shvo2022ResolvingMA}. This process, if repeated until no contradictions are found, results in a coherent mental model, and the entire process is designated as explanation. Since each step in the process is conditioned on explainee's current mental model and the contradictions that are observed by the previous iteration---explanation in its ideal form is \textit{interactive}~\cite{DBLP:journals/corr/abs-1804-09299,miller19-social,DBLP:journals/tvcg/GehrmannSKPR20,DBLP:journals/corr/abs-2109-07869}.


\section{Composition of an Explanatory Narrative} \label{sec:folk-psychology-model}


Assume that our goal now is to enable explainees to establish coherent mental models of an AI's behavior, according to the definition in \Cref{sec:faithfulness}. In this section we look at precedence in theory of mind research (\Cref{sec:anthropomorphism-intent}) to construct an explanatory narrative that can aid this goal (\Cref{subsec:folk-concepts}).

\subsection{Anthropomorphic Bias and Perceived Intentionality} \label{sec:anthropomorphism-intent}

\begin{figure*}[t]
\centering
  \includegraphics[width=0.77\textwidth]{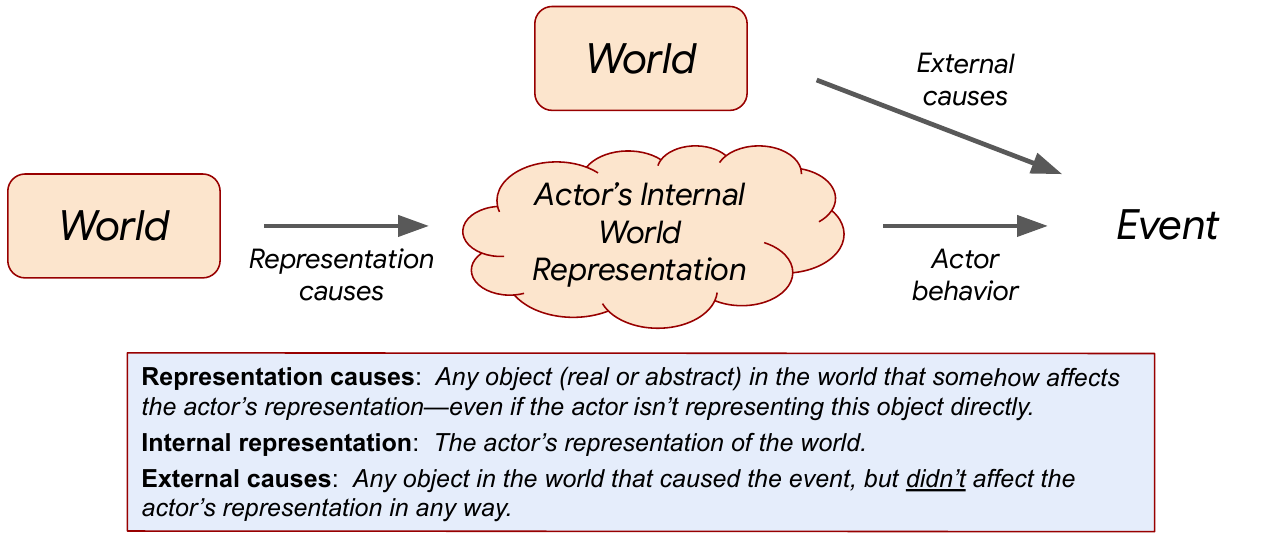}
  \caption{Folk concepts of behavior (adapted from~\citet{malle2003}). Research shows that humans understand and explain events along these concepts. See \Cref{subsec:folk-concepts} for description and examples.}
  \label{fig:folkconcepts}
\end{figure*}

\textit{Intentionality} is a central concept in models of folk theory of mind~\cite{karniol1978-children-intent,Knobe1997-folk-intentionality,burra-folk-intention}: It refers to the power of mind to internally represent things about the world.\footnote{Note that this sense of the word ``intentionality'' in philosophy differs from the colloquial sense of intent in English.} When we comprehend explanations about events, we intuitively do so with respect to ``actors'' which hold internal representations, and whose behaviors had a causal role on the event (\Cref{fig:folkconcepts}). 

The explainee's assumptions about explained events are potentially biased with respect to how humans think and behave: If there is an actor in the event's history, we may potentially understand this actor (human or not) by imagining how we may have acted in the actor's circumstances, implicitly assigning a mental representation to the actor~\cite{culley2013note}. 

When the actor is not human, we refer to this as \textit{anthropomorphic bias}. This bias is widespread and common~\cite{Dacey2017-anthropomorphic-bias,johnson2018-anthropomorphic-bias}.
For example,~\citet{Heider1944AnES} found that humans attribute human-like behavior to simple moving shapes. Regardless of the nature of extent of this bias, if the explainee can view the AI as an actor capable of holding internal representation, explanations of events concerning the AI must account for this fact in some way---either to suppress this bias, or to clarify it.

The bias in attributing an internal representation to AI processes is prevalent in the general public and even domain and AI experts~\cite{darling2015-anthropomorphic-framing-robots,Salles2020AnthropomorphismIA}. For example,~\citet{ehsan2021-ai-experts-explanation-perception} found that AI experts (computer-science students of AI curriculum) and non-experts alike, through explanations, attribute modes of human-like power of mind to AI behavior, even (though less so) when the explanations do not contain explicit information about justification behind the AI's decisions, and the effect is stronger when the explanation is given in natural language. Additionally, \textit{concept explanations}~\cite<TCAV,>{kim2018-tcav} are an explicit attribution of symbolic representation to AI (\Cref{subsec:concept-function}), and \textit{natural-language explanations}~\cite{narang-et-al-2020-wt5,DBLP:journals/corr/abs-2102-12060} attempt to give AI a human voice (\Cref{subsec:ng-rationales-function}). Even the act of \textit{text marking}, a common explanation format in XAI, can be interpreted with an anthropomorphized lens~\cite{Marzouk2018TextMA,DBLP:journals/tacl/JacoviG21}.\footnote{\citet{Marzouk2018TextMA} note many possible attributions of intentionality to text marking: Marking ``easy to forget'' text, marking definitions, marking unclear text to investigate later, summarizing text, marking text contradictory to personal belief, and so on. When reading marked text, the perception of how this marking came to be influences how it is understood.} Finally, AI researchers and developers are susceptible to using anthromopomorphic rhetoric, as well~\cite{watson2020-anthropomorphism-rhetoric}.

\noindentparagraph{On mitigating anthropomorphic bias.} The attribution of human-like internal representation to AI as a result of anthropomorphic bias is implicit, possibly of subconscious habit, and is therefore potentially damaging to the utility of AI explanations~\cite{ehsan2021-ai-experts-explanation-perception,Hartzog2015UNFAIRAD}.\footnote{Whether humans can be ``correct'' in attributing mental states to AI at all is a matter of philosophical debate, but nevertheless there is sufficient evidence that humans do make this attribution often \cite{uncanny-believers}. We argue that explaining AI behavior successfully, so that explainees' mental models are coherent, is a goal which is independent of the discussion of whether these mental models are philosophically ``correct'' or not, if AI is to be useful in society.} There are three possible methods of mitigating this danger: (1) To adapt to the bias by understanding the perceived power of mind, and taking action on AI design to accommodate it~\cite{Zlotowski2015AnthropomorphismOA}; (2) to control the perception of power of mind by taking action to properly communicate the AI's capabilities~\cite{darling2015-anthropomorphic-framing-robots}; or (3) to remove it entirely by communicating to the explainee the lack power of mind in an AI (see e.g., scientific explanation of natural phenomena, such as explaining how planes fly, or explaining how tools work)~\cite{epley2007-anthropomorphism-three-factor}.\footnote{In particular, human-robot interaction research discusses all three methods with respect to robots: For example,~\citet{10.1145/3319502.3374839} conduct a user study controlling for anthropomorphic rhetoric in human-robot interaction, through personification, and feedback such as apology or indifference; finding significant effect on trust. \citet{darling2015-anthropomorphic-framing-robots} discusses anthropomorphic framing of robots, and argue for both beneficial and detrimental aspects of anthropomorphism, and aspects that control it (framing robots as tools or as companions). \label{foot:robots-anthromorphism}}


\subsection{Categories of Folk Concepts of Behavior} \label{subsec:folk-concepts}



\Cref{fig:folkconcepts} outlines a categorization of causes that, according to empirical evidence, humans comprehend intuitively~\cite{karniol1978-children-intent,Knobe1997-folk-intentionality,malle2003,burra-folk-intention}: Representation causes, external causes, and internal representations, with respect to a contrast case. We describe each of these concepts and demonstrate their use in a running example.\footnote{Terms and categorizations in this section are simplified slightly from their philosophy counterparts, to reduce the barrier of entry for the AI audience.}



\begin{figure*}[t]
\centering
  \includegraphics[width=0.85\textwidth]{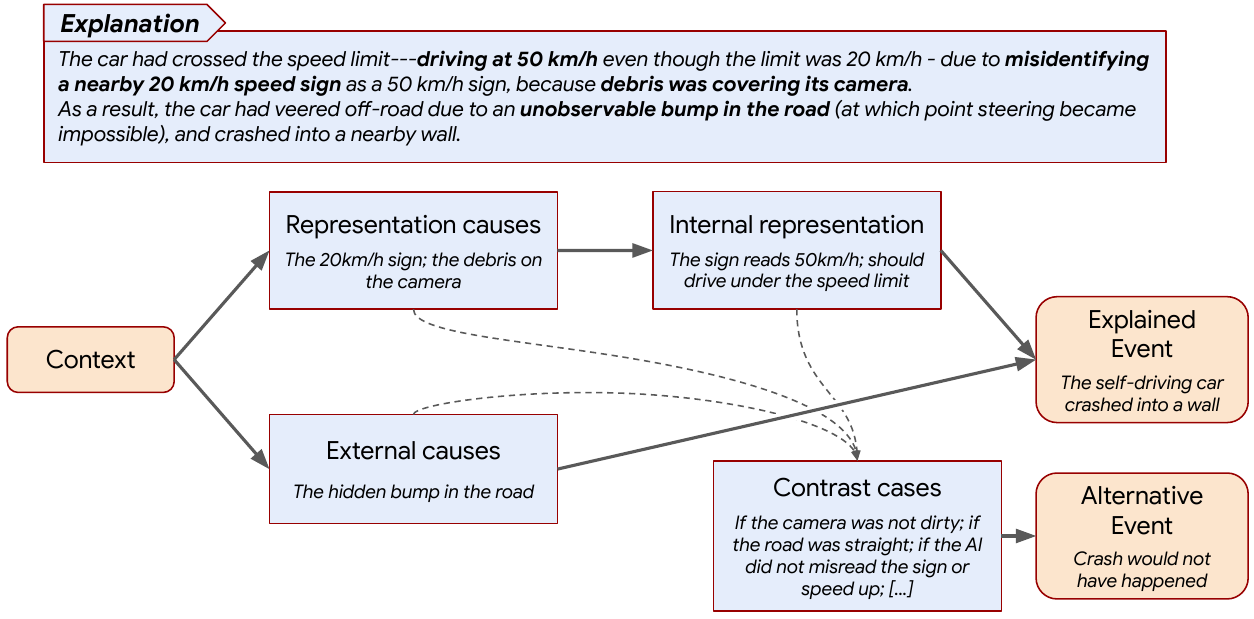}
  \caption{Schematic of the \textit{self-driving car} example in the categorization the explanatory narrative discussed in \Cref{sec:folk-psychology-model}.}
  \label{fig:malle03}
\end{figure*}

\noindentparagraph{\textbf{Running example (\textit{self-driving car},} \Cref{fig:malle03}).} Consider a self-driving car that was involved in an accident: The car drove into a wall. An explanation is provided: The car had crossed the speed limit---driving at $50\ km/h$ even though the limit was $20\ km/h$---due to misidentifying a nearby $20\ km/h$ speed sign as a $50\ km/h$ sign, because debris was covering its camera. As a result, the car had veered off-road due to an unobservable bump in the road (at which point steering became impossible),
and crashed into a nearby wall. 

Supposing that the explanation is ``true'', we assume that a human explainee considers the AI software in the car as an actor, and that they consider the explanation satisfactory. We will highlight one possible mental model that could manifest for this example.

\subsubsection{Internal Representation} Internal representation refers to how the actor subjectively represents the world (independently from the actual world). E.g., ``The man robbed the bank because \textit{he needed money}''.\footnote{
There are multiple models of mental states in philosophy, the most common and simplistic being that of collections of beliefs and desires; additional models include values, emotions, thoughts, outcome-beliefs and ability-beliefs~\cite{heider1958-abilitybelief-outcomebelief} among others~\cite{malle2003,Andrews2006-ANDTFO}. It can be argued that promoting the attribution of beliefs and desires to automated processes encourages excessive anthropomorphism of machines. In this work, we discuss the attribution of internal representation only to the extent of evidence that it occurs, without adopting a specific definition for what ``internal representation'' can refer to, as this is an active area of debate~\cite{uncanny-believers}.} 

\textit{Running example:} The explainee may understand that the car's software internally represents the sign as a sign of a $50\ km/h$ speed limit, and is representing the rule to drive at or below the speed limit.

\subsubsection{{Representation Causes}} Representation causes refer to causes in the world that causally affected the actors' internal representation in some way (i.e., if we intervene on the representation cause, the actor's internal representation would change).\footnote{Representation causes and the representation itself have different roles in communicating information about the actor's behavior; for example,~\citet{brem-rips-2000} found that evidence (objective causes) is considered more explanatory among more knowledgeable explainees in legal settings, in comparison to explicitly explaining subjective internal representation directly, than among explainees with less expertise.} E.g., ``The man robbed a bank because he needed money to treat \textit{an illness}''---if the illness did not exist, the man would not need the money (``needing money'' being an internal representation).

\textit{Running example:} The sign of $20\ km/h$, the camera, and the debris, are all objective causes of representation, as they provide causal history to how the AI represented the sign. In other words, if one of these factors was different in some way (e.g., no debris on the camera, or sign of a roundabout ahead), our explainee would potentially expect the AI's internal representation to be different as well.

\subsubsection{{External Causes}} External causes are the objective causes in the world which are \textit{unrelated} to an actor's internal representation. E.g., ``the man successfully robbed the bank because \textit{the security alarm was faulty}''---whether the security enabled the robbery or not, has no effect on the man's intentionality to rob the bank.

\textit{Running example:} Our explainee may comprehend the unobserved bump in the road as an external cause: In the explainee's mental model of the accident, regardless of whether the road bump existed or not, the AI's \textit{internal representation} would not change---the car's AI would still misidentify the sign, and drive at $50\ km/h$. However, the final event would change---the accident would not have happened---which means that the hidden bump did have a causal effect on the accident without affecting the AI's representation of the world. 

\subsubsection{\textbf{Contrast Cases}} \label{subsec:counterfactual-bifactual}  Explanations, as a function of mental models, are widely accepted to be contrastive~\cite{lugg_1983,lipton_1990-contrastive,hilton1990-causalexplanation}. This is due to the limit of cognitive load of humans to process ``complete'' explanations~\cite{Lewis1986-contrastive,miller18-contrastive}, so the explanation is simplified by contrasting the event against another event of similar context~\cite{petri06-contrastive-explanandum}. 
As such, all of the previous categories of folk concepts (\textit{internal representation}, \textit{representation causes}, and \textit{external causes}) can be comprehended with respect to the contrast case that intervenes on them. 

We can make a distinction between ``bifactual'' and ``counterfactual'' contrast cases: Bifactual being an event which occurred in reality (answering ``why did P happen in context A, while Q happened in context B?''), and counterfactual being a theoretical-fictional event (answering ``why did P happen in context A instead of Q?'')~\cite{miller18-contrastive}. 

\textit{Running example:} The debris on the camera, being a representation cause, implies a counterfactual reality: ``Had the camera not been dirty, the car would have not misidentified the sign.'' The same information can be given via a bifactual instead: ``Last week, the car had driven on the same road with a clean camera, at $20\ km/h$, and the accident did not occur.''

\subsection{Implications}

The categorization of folk concepts of behavior has two relevant implications for constructing an explanatory narrative:


\noindentparagraph{Representation causes and representation form a causal chain, such that explanations without \textit{both} components are more difficult to understand.} Explaining a representation cause without the resulting internal representation may force the explainee to assume what the representation is; explaining the internal representation without the causes that led to it may force the explainee to assume what those causes were. We explore this in \Cref{sec:analysis-xai-methods}.

\noindentparagraph{The explainee may make incorrect generalizing assumptions by assuming missing components.} The step of interpreting \textit{representation causes} into \textit{representation} by the explainee serves to apply more general rules that conform to the causal history coherently~\cite{murphy1985}: We attribute an internal representation to the actor based on our knowledge of what representation \textit{we} may have had in a similar context~\cite{Andrews2006-ANDTFO}. If the representation is hallucinated or misunderstood, this attribution may be wrong, and thus incoherent~\cite{DBLP:conf/chi/Lewis86}. As~\citet{DBLP:journals/jasss/NowakRB13} explain, mental models of abstract, non-linear processes happening in complex systems are almost impossible to construct solely using individual cognitive capabilities.



\section{The Narratives of AI Explanation Methods} \label{sec:analysis-xai-methods}

Analyses of XAI methods often focus on their ability to satisfy heuristics of what explanation methods should do, and conclude that they are fragile~\cite{Kindermans2019,NEURIPS2019_fe4b8556,jacovi2020-faithfully}. But it remains unclear what exactly is the point of failure, in terms of the potential explainee's mental model, and the contradictions between it and observed behavior.

Using the actor-centric framework developed thus far (\Cref{subsec:folk-concepts}), we are now able to diagnose a given XAI method for potential contradictions between what the method communicates about model behavior, and the mental model of the explainee from which they extrapolate AI behavior. 

This section is a case study of such diagnoses of four common types of AI explanation. Each diagnosis follows a general structure: (1) Description of the mechanism; (2) The possible information it communicates, in the language of folk concepts of behavior; (3) An illustrative example of the resulting perceived narrative; (4) Diagnosis of potential failure modes. We provide an overview in \Cref{tab:diagnosis-summary}.






\noindentparagraph{\textbf{Assumptions.}} (A) \textit{On internal representation:} We assume that the AIs in the case studies are perceived as actors capable of holding internal representations (see \Cref{sec:anthropomorphism-intent}). 
(B) {\textit{On correct explanation:}} 
We are not concerned in this section with whether the explanations are ``faithfully'' describing the model (see \Cref{sec:faithfulness}), but only in how a fictional explainee may comprehend them. 
(C) {\textit{On interactive explanation:}} For demonstration we assume a single iteration for explanation to surface possible contradictions in the scope of the iteration. This is not to say that the explanation is ``forfeit'' once contradictions surface, but that \textit{additional iterations are required} to re-establish coherence. Each iteration of explanation is a direct result of the previous iteration's mental model, which makes interactivity indispensable for its implementation. As a general approach, in all of the following cases, once an issue is found---the hypothesis could be adjusted by exploring explanations for additional examples.

\begin{table}[t]
    \centering
    \ra{1.2}
    \caption{Summary
    of \Cref{sec:analysis-xai-methods} and application to various explanation methods. ($*$) In standard neural networks, the contrast case is explicit for continuous-space inputs (vision, speech) but implicit for embedded inputs (discrete sequences, natural language). 
    }
    \resizebox{0.99\linewidth}{!}{
    \begin{tabular}{>{\raggedright}p{3.5cm}>{\raggedright}p{2.3cm}>{\raggedright}p{2.3cm}p{11cm}}
        \toprule \textbf{Mechanism} & \textbf{Explicit components} & \textbf{Missing components} & \textbf{Possible failure (e.g.)} \\
        \midrule Similar training examples (\S\ref{subsec:influence-functions}) & Contrast case & Repr. cause, representation & \textit{Contradiction with perceived repr. cause \textbf{(A)}} \newline Explainee will look for similarities (repr. causes) between examples that are important to model behavior.  \newline \textit{Contradiction:} The hypothesized similarities are unimportant to model behavior, s.t. model behavior and expected model behavior will be different on additional examples which share the similarities. \\
        Influence functions~\cite{koh2017-influence-functions} (\S\ref{subsec:influence-functions}) & Contrast case, repr. cause & Representation & \textit{Contradiction with perceived representation \textbf{(B)}} \newline Explainee will assume that the model learned to ``represent and use'' some property (repr. cause) in the influential example, where the property is a shared characteristic between the real and influential example, despite a different model representation. \newline
            \textit{Contradiction:} Model behavior will differ from expectation on additional examples with the property, due to the different internal representation. \\
        Feature attribution\footnotemark \ (\S\ref{subsec:feature-attribution}) & Contrast case$^{(*)}$, repr. cause & Contrast case$^{(*)}$, representation &  \textit{Contradiction with perceived representation \textbf{(B)}} \newline Explainee may assume that the model is interpreting some word in the input (representation) in a specific context (e.g., using a gender pronoun to signal gender) while the model is using it for something else (e.g., the co-referred noun of the pronoun). \newline 
        \textit{Contradiction:} Model behavior will differ from expectation on examples that share the same repr. cause (the same gender pronoun), but differ in representation (the entity that the pronoun is referring to). \\
        TCAV~\cite{kim2018-tcav}, MDL probing~\cite{DBLP:conf/emnlp/VoitaT20} (\S\ref{subsec:concept-function}) & Representation & Contrast case, repr. cause &  \textit{Contradiction with perceived context \textbf{(C)}} \newline  Model recognizes that some property (e.g., striped fur) was in the image, but counterfactual is missing: Explainee may assume ``striped fur rather than mono-color fur'', but the real contrast case may be ``striped fur rather than dotted fur''. \newline 
        \textit{Contradiction:} Model behavior will differ from expectation on examples which share properties with the hypothesized counterfactual (e.g., mono-color fur examples). \\
        Amnesic Probing~\cite{elazar2021-amnesic-probing}, CausalM~\cite{feder2021-causalm} (\S\ref{subsec:concept-function}) & Contrast case, representation & Repr. cause &  \textit{Contradiction with perceived repr. cause \textbf{(A)}} \newline 
        The explainee may assume that some part of the example caused the representation (e.g., whiskers in the image and the model recognizing whiskers), while the representation is based on a different repr. cause. \newline
        \textit{Contradiction:} Model behavior will differ on examples which share the real repr. cause (e.g., blades of grass), but not the perceived repr. cause (e.g., whiskers). \\
        WT5 rationalization (\S\ref{subsec:ng-rationales-function}) & Representation & Contrast case, repr. cause &  As the function of this mechanism is the same as \textit{concept attribution}, so are its failures.  \\   
        \bottomrule
    \end{tabular} }
    \label{tab:diagnosis-summary}
\end{table}

\footnotetext{Gradients \cite{li-etal-2016-visualizing,DBLP:journals/ijcv/SelvarajuCDVPB20}, SmoothGrad~\cite{DBLP:journals/corr/SmilkovTKVW17}, LIME~\cite{DBLP:journals/corr/RibeiroSG16}, SHAP~\cite{DBLP:conf/nips/LundbergL17}, inter alia.}

\subsection{Training Data Attribution} \label{subsec:influence-functions}

\noindentparagraph{\textbf{Mechanism.}} A class of methods for supervised AI models attempt to attribute the examples in the training data which ``influenced'' a particular decision. Influence functions approximate the effect of removing an example from the training data on the loss of the explained example~\cite{koh2017-influence-functions,DBLP:conf/acl/HanWT20}; Cook's distance measures the change in prediction for an example for linear regression models by removing a training example from training and re-training the model~\cite{cook1977}.

\noindentparagraph{\textbf{Folk concepts.}} The influential examples produced by training data attribution methods can be interpreted as \textit{representation causes}: Communicating what influences the AI's representation of the input. But interestingly, since influential examples do not communicate the contrast case, they can be perceived in two different ways: (1) As \textit{bifactual}, i.e., ``explanation by example.'' The influential example is simply another related instance of behavior; (2) or as a \textit{counterfactual},
a contrast case in which if the influential example was not part of the AI's training, its loss function on the current example would have changed. The two different perspectives can potentially change how the explainee will understand the explanation.



\begin{figure*}[t]
\centering
  \includegraphics[width=0.9\linewidth]{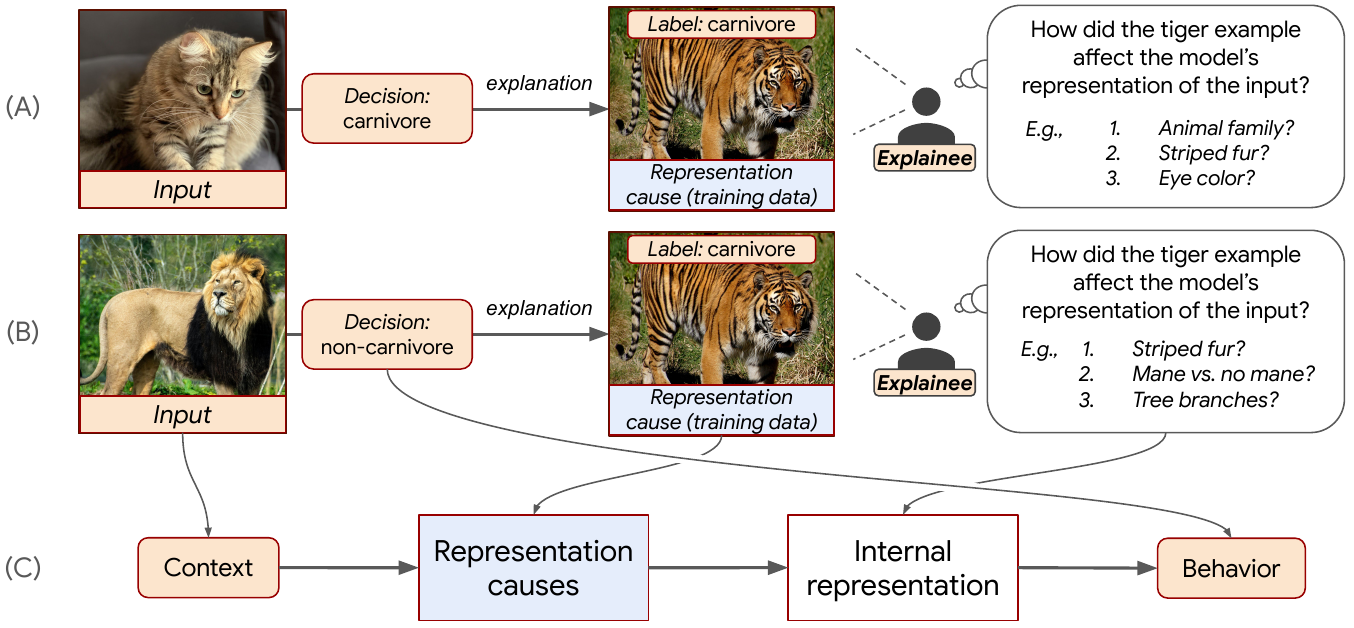}
  \caption{Demonstration of the explanatory narrative of influential examples (\Cref{subsec:influence-functions}). The ``influential example'' explanation is the same in both contexts, yet only constitutes as a \textit{representation cause}, and can be interpreted in very different ways by imagining the missing explanation for \textit{how the model represents the cause}. (C) shows the incomplete mapping to the folk concepts that compose the explanatory narrative.
}
  \label{fig:test}
\end{figure*}

\noindentparagraph{\textbf{Demonstration (\textit{carnivorism prediction}).}} 
Consider the case of a classifier that classifies whether a given image of an animal is a carnivore or a non-carnivore. The AI model takes an image of a cat and outputs the decision that it is a carnivore (\Cref{fig:test}a). An influence function-based method provides an explanation as a training data image of a tiger which influenced the prediction.

If interpreted as a counterfactual, an explainee may understand that the model is making a generalization for carnivores based on shared characteristics between the images (e.g., that they belong to the carnivorous felidae family; or that a striped fur is indicative of carnivorism).\footnote{Of course, it is also possible to interpret the explanation differently. Assume this interpretation for the sake of demonstration.}

If interpreted as a bifactual, however, it is possible to formulate a different mental model from the \textit{same} explanation in a different context. Suppose that the model erroneously categorizes a picture of a lion as a non-carnivore, but the explanation given is the same training-data image of a tiger (\Cref{fig:test}b). 
The meaning of the explanation could now be perceived as an answer to the question: \textit{Why did the model decide that the lion is not a carnivore---while the tiger is?}

\noindentparagraph{\textbf{Potential failure (\textit{implicit internal representation}).}} The reason for the two different interpretations for the same explanation, described above, is that the explanation does not communicate what the model is representing in the relationship between the two images. Therefore, the explainee is free to make assumptions about this representation, which may be correct or incorrect. An incorrect assumption would lead to a contradiction when the model behaves differently from expected in the future.\footnote{
This issue is further exacerbated by the fact that many AI systems, in particular neural models which are explained by influence functions, do not possess a symbolic internal representation system. This complicates the exercise of making ``correct'' assumptions about the models' internal representations.}


\subsection{Feature Attribution} \label{subsec:feature-attribution}


\noindentparagraph{\textbf{Mechanism.}} Feature attribution methods, whether discrete (e.g., LIME,~\cite{DBLP:journals/corr/RibeiroSG16} and erasure~\cite{DBLP:journals/corr/ArrasHMMS16a,feng-etal-2018-pathologies}) or continuous (e.g., gradient-based~\cite{DBLP:journals/corr/SimonyanVZ13} and attention flow~\cite{abnar2020-attention-flow,ethayarajh2021-attention-flow-shapley}), derive which portions of the input have influence on the AI's behavior by intervening on (perturbing) the input in some systematic way and observing behavior changes.

\noindentparagraph{\textbf{Folk concepts.}} Feature attribution methods provide \textit{representation causes}---regions in the input to the model (e.g., phrases in text or pixels in image) that especially influenced how the model represents the input. However, they traditionally do not provide information on how the models represent those regions internally.

\noindentparagraph{Demonstration (\textit{restaurant review}).} Suppose that in a \textit{sentiment classification} task, a classifying model predicts the binary sentiment polarity of a restaurant review:
\begin{quote}
    Best \underline{Mexican} I’ve ever had!  $\longrightarrow$ \textit{positive}
\end{quote}
Where the underlined text is the explanation. This explanation is likely to be interpreted as a representation cause: The claim is that if this part of the input changed, then the classifier's internal representation of the input will change significantly, and therefore the decision would also change. 

\noindentparagraph{\textbf{Potential failure (\textit{implicit contrast case and internal representation}).}}
The explanation is missing what the classifier is representing in the attributed phrase, and what type of intervention on this phrase would change the classifier's representation. Therefore, the explainee is at risk of assuming what these missing are. For illustration, below are two possible assumptions about this representation and contrast case:
\begin{quote}
    (1) Best \underline{Indian} I’ve ever had! \textit{(country identity)} \\
    (2) Best \underline{fish} I’ve ever had! \textit{(food category)}
\end{quote}
This ambiguity is a potential point of fragility in the explainee's comprehension of the model's behavior. Without additional clarity on these folk concepts, the explainee may assume one of the options, and discover a contradiction if the assumption is incorrect.

\subsection{Concept Attribution} \label{subsec:concept-function}

\noindentparagraph{\textbf{Mechanism.}} A class of XAI methods attempt to characterize which human-interpretable abstractions (concepts) are represented by, and used in, the AI model's reasoning process. In this area, probing methods~\cite{DBLP:conf/iclr/AdiKBLG17,DBLP:conf/acl/BaroniBLKC18} characterize what is encoded in the model's representation, while TCAV~\cite{kim2018-tcav}, MDL probing~\cite{DBLP:conf/emnlp/VoitaT20}, amnesic probing~\cite{elazar2021-amnesic-probing}, causal mediation analysis~\cite{vig2020-causal-mediation-nlp}, causal abstractions~\cite{geiger2021-causal-abstractions}, inter alia, provide more insight on the role of the concepts in model behavior.


\noindentparagraph{\textbf{Folk concepts.}} Concept attribution methods map the AI model's internal representation of the input into human-interpretable concepts, therefore they attempt to communicate internal representation.
Importantly, whether a concept is detected in the internal representation does not entail whether the concept really does or does not exist in the context. 

\begin{figure*}[t]
\centering
  \includegraphics[width=0.95\linewidth]{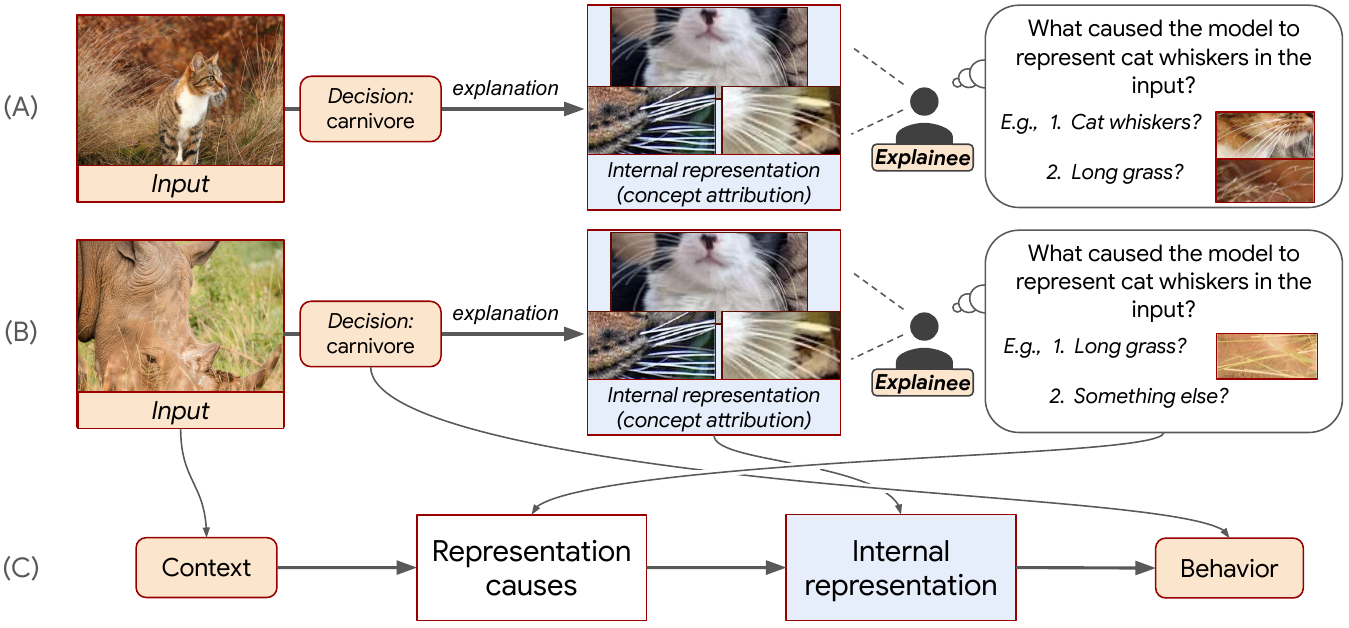}
  \caption{Example for concept explanations (\Cref{subsec:concept-function}). The explainee may hallucinate the cause of the attributed concept to be the whiskers in the image (or any particular object in the image), even though this is not part of the explanation: The explanation only communicated the internal representation of the model, but not what could have affected this representation. (C) shows the incomplete mapping to the folk concepts that compose the explanatory narrative.}
  \label{fig:concept-whiskers}
\end{figure*}

\noindentparagraph{\textbf{Demonstration (\textit{whiskers attribution}).}} Suppose that a model decision, classifying a cat image as a carnivore, has been attributed with the ``cat whiskers'' concept (\Cref{fig:concept-whiskers}a). The concept is commonly defined as a set of samples that contain the concept.


\noindentparagraph{\textbf{Potential failure 1 (\textit{implicit representation causes}).}} Current methods of concept explanations communicate internal representation without its causes. This means, for example, that the explainee may understand that the model represents that the image of a cat has whiskers, but not necessarily what caused this.

This is fragile point in the explanatory narrative: The explainee may make an assumption about what caused the representation of the concept, and this assumption may not be true. For example, if the image indeed has a cat with whiskers, the explainee may assume that the model's representation of the whiskers concept is caused by the whiskers in the image---when in reality, perhaps the model mistook blades of long grass in the background of the image for whiskers. 
The assumption can cause a failure of coherence if the model behaved similarly on other images which do not have whiskers, but do have similar blades of grass (\Cref{fig:concept-whiskers}b).

\noindentparagraph{\textbf{Potential failure 2 (\textit{implicit contrast case}).}} In the case of classic \textit{probing} methods which communicate whether a concept is being represented by the model, it is possible that this representation is not a cause of the model's final decision (i.e., it does not explain the decision). This is because the counterfactual case, where the concept is absent, is not part of the explanation. 

This has been a subject of recent criticism for probing methods, on the basis of ``correlation does not equal causation''---where although probing methods infer that the model represents some concept, no guarantee is given if the model actually uses this concept to make its decisions~\cite{DBLP:conf/emnlp/TamkinSGG20,DBLP:conf/blackboxnlp/GeigerRP20,DBLP:conf/eacl/RavichanderBH21}. This has led to the development of a causally-informed class of methods~\cite{vig2020-causal-mediation-nlp,feder2021-causalm,geiger2021-causal-abstractions} which provide a stronger guarantee that causality is correctly attributed. This can be accomplished, for example, by showing that the model changes its decision if it ceases to recognize the concept via a counterfactual~\cite{elazar2021-amnesic-probing,feder2021-causalm}.


\subsection{Natural-language Generation (a.k.a. Abstractive Rationales)} \label{subsec:ng-rationales-function}


\noindentparagraph{\textbf{Mechanism.}} 
Models generating ``rationalizations'' as natural-language explanations~\cite{ehsan-et-al-2018-rationalization,DBLP:journals/corr/abs-2010-12762,narang-et-al-2020-wt5} learn from human-written explanations to produce a natural text from the AI model's hidden representation, attempting to justify their actions, inspired by the way that a human would explain their own behavior~\cite{DBLP:journals/corr/abs-2102-12060}. 

\noindentparagraph{\textbf{Folk concepts.}} This class of explanations attempts to communicate what the model is representing in natural language, therefore they communicate the model's \textit{internal representation}. Note that this is a very similar narrative function to \textit{concept attribution} (\Cref{subsec:concept-function}). The medium of natural-language communication may reinforce anthropomorphic bias in comparison to other mediums~\cite{ehsan2021-ai-experts-explanation-perception}. 

\noindentparagraph{\textbf{Demonstration.}} Continuing the \textit{whiskers attribution} example from \Cref{fig:concept-whiskers}, such a model may generate the explanation: ``\textit{Because it has whiskers}'', ``\textit{because it has stripes}'' or even ``\textit{because it eats meat}'' as a rationalization. 

\noindentparagraph{\textbf{Potential failures.}} Natural language rationalization communicates the same folk concepts of behavior as concept attribution, therefore it shares the same potential coherence failures (for example, implicit \textit{representation causes}), despite these two methodologies having very different underlying technologies.

\section{Towards Successful Explanations} \label{sec:diagnosis-conclusion}

\begin{table}[t]
    \centering
    \ra{1.3}
    \caption{Various, seemingly different, XAI methods may share the same failures according to our abstraction (\Cref{sec:diagnosis-conclusion}).}
    \resizebox{0.92\linewidth}{!}{ 
    \begin{tabular}{cl>{\raggedright}p{4.2cm}l}
        \toprule \multicolumn{2}{l}{\textbf{Incompleteness type}} & \textbf{Potential 
        failure} & \textbf{Mechanisms}   \\
        \midrule  \textbf{(A)} & Missing representation cause & Contradiction with perceived representation causes & Bifactual training examples, concept attribution \\
        \textbf{(B)} & Missing internal representation & Contradiction with perceived representation & Training data attribution, feature attribution \\
        \textbf{(C)} & Missing contrast case & Contradiction with perceived contrast case & Feature attribution, concept attribution  \\ \bottomrule
    \end{tabular} }
    \label{tab:failure-types}
\end{table}

\begin{figure*}[!t]
\centering
  \includegraphics[width=0.9\linewidth]{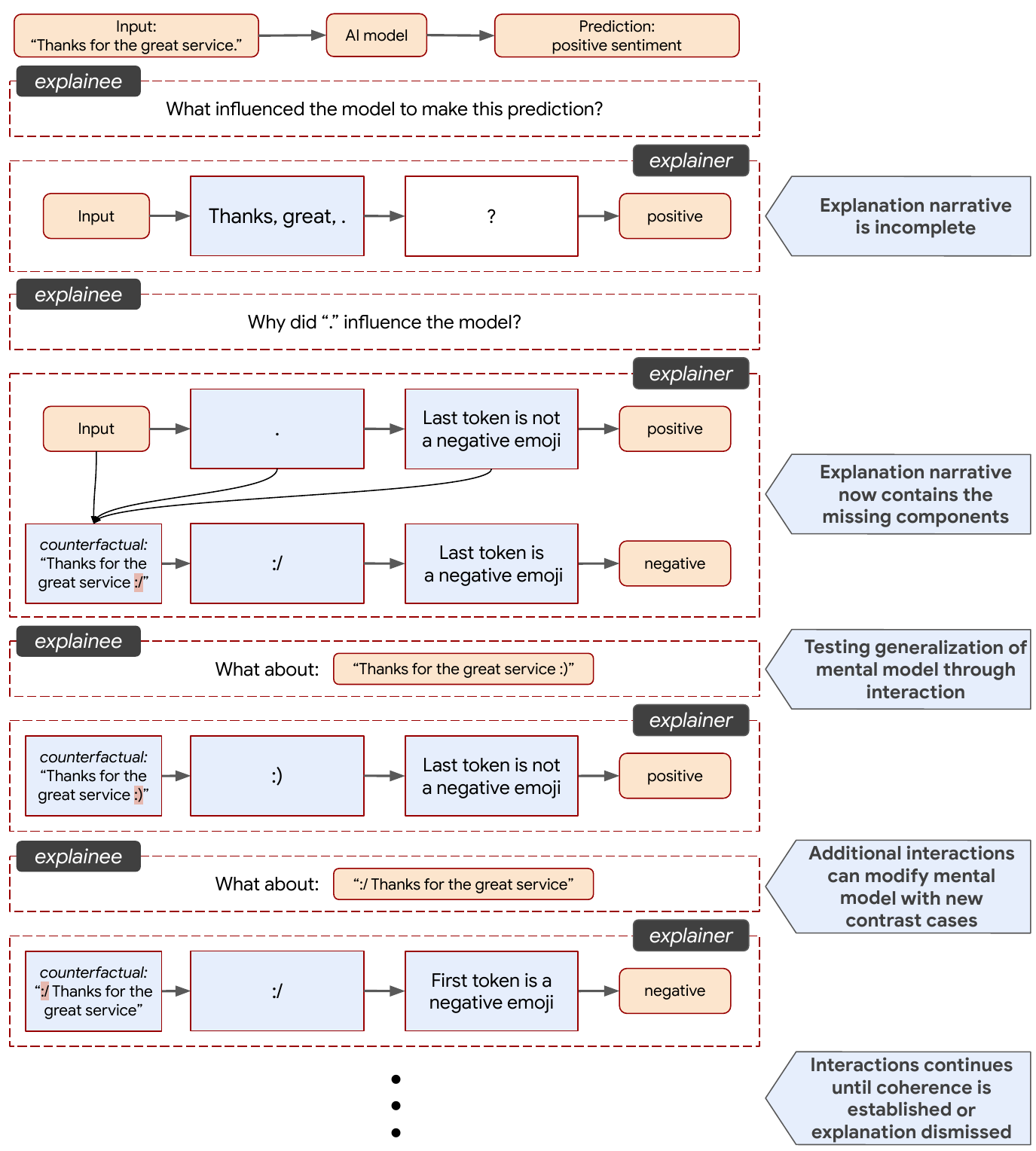}
  \caption{Illustrative example of how interactive interrogation and a complete explanatory narrative can serve as modes of explanation (\Cref{sec:diagnosis-conclusion}).}
  \label{fig:complete-interactive}
\end{figure*}

We summarize the main implications from the analysis thus far:

\noindentparagraph{Explanations should establish an explanatory narrative which explicitly communicates all relevant folk concepts of behavior.} 
The underlying root issue in all potential failures discussed in \Cref{sec:analysis-xai-methods} is \textit{an under-specification of the AI process by the explanation} (\Cref{tab:failure-types}). Unaccounted components of the explanatory narrative are at risk of being ``filled in'' by the explainee through potentially incorrect assumptions, leading to contradictions. The explanatory narrative we propose is as follows: ``Something'' in the context (input data, training data, or algorithm; \textit{representation causes}) caused the AI to represent ``something'' (\textit{internal representation}) which affected the explained outcome; intervening on the representation causes will change the representation, ultimately changing the outcome (\textit{contrast case}). Additional relevant causes which had no effect on the AI's representation, but nevertheless affected the outcome (\textit{external causes}) should be explicitly marked as such, with additional contrast cases. 
See an illustrative example in  \Cref{fig:complete-interactive}.


\noindentparagraph{Explanations should use interactivity to resolve contradictions.} In this work we regard something as ``successfully'' explained if the explainee can establish a coherent mental model, without observable contradictions to it. But explanations do not necessarily need to accomplish this in ``one shot'', as humans naturally use interactivity to adjust incoherent mental models. 
Therefore, we stand to make breakthroughs in successfully explaining AI not only by improving the explanatory narrative that the explanation communicate, but also by allowing the explainee to test their hypothesis via interactive interrogation of the AI~\cite{DBLP:journals/corr/abs-1907-10739,gehrmann2020human,DBLP:journals/jair/KrarupKMLC021}.

\noindentparagraph{Additional research is required on \textit{explainee profiling}~\cite{gerhard2000-user-modeling,addie2005-user-modeling} to characterize how different explainees may construct mental models differently.} The definition of a successful explanation, as a function of a coherent mental model, is a definition that involves the explainee. In order to understand the mental model of the explainee, we must establish who the explainee is, and what prior knowledge they may leverage in their assumptions. Currently, explainee profiling in XAI is often limited to familiarity with AI technology or expertise at the end-task~\cite<``AI experts/novices'', ``domain experts'', ``data scientists''; e.g.,>{strobelt2017lstmvis,DBLP:journals/tvcg/HohmanKPC19,10.1145/3313831.3376219,ehsan2021-ai-experts-explanation-perception}, but additional research may uncover other important properties of user models, such as cognitive or social properties.

\section{Discussion}

\noindentparagraph{On scientific explanation and argumentation.} As mentioned, this work is concerned with communicating information about the process that led to AI behavior, rather than justifying this behavior or verifying its correctness (argumentation) \cite{DBLP:journals/corr/abs-2010-12896,fok2023search,miller2023explainable,langley2019varieties}. The process of explaining behavior is partially discussed by literature on scientific explanation \cite{sep-scientific-explanation}, though this literature is generally more concerned with the specifics of deriving causal insights about processes, while this work is concerned with how (or what it would require) to \textit{properly communicate} these derived insights to humans, specifically in the case of AI behavior. 

\noindentparagraph{On external causes.} External causes are one of the central building blocks of an explanatory narrative explored in this work. However, the explanation methods and failure cases discussed in \Cref{sec:analysis-xai-methods} do not mention external causes. Why is this the case? The AI and XAI utilities discussed here pertain to the AI mechanism itself, outside of a context in which external factors can have an effect. For example, in a robotics use-case, external hardware factors (such as water damage or wear) can affect behavior, and thus have a place in an explanatory narrative. However, such factors are detached from XAI methods that only attempt to explain the AI component.

\section{Conclusion}

This work identifies two different perspectives of explanation: (1) What the explanation method is communicating about the AI behavior; (2) what the explainee actually comprehends about AI behavior from the explanation. We find that the explainee may derive incorrect generalizing rules about AI behavior, causing a mismatch between (1) and (2), if the explanation is unintuitive or insufficient. 

Erroneous generalizing assumptions will cause contradictions to manifest between \textit{additional} AI behavior and the explainee's mental model. In the event of observed contradictions, we say that the mental model is incoherent, and that coherency is a primary attribute of good explanation. Successfully explaining without contradictions does not necessarily require a ``perfect'' initial explanation, since contradictions can be resolved via interactive interrogation of AI behavior, iteratively adjusting the mental model until it is coherent.


We apply this framework to a variety of XAI methods, and find that contradictions systematically arise from missing information in the explanation (in terms of how humans comprehend explanations: Through representation causes, internal representation, external causes and a contrast case). This provides us with a path forward towards the design of XAI methods that can be said to provide coherent explanation, specifically by being complete and interactive.

\noindentparagraph{Extensions and future work.}
We note additional exceptionally multi-disciplinary research directions related to successful explanation:
\begin{enumerate*}
    \item How to communicate lack of power of mind in what society considers as AI, or ``intelligent'' automated processes. As noted in \Cref{foot:robots-anthromorphism}, this question is discussed in \textit{human-robot interaction}, but remains an open question in other settings.
    \item Characterizing the budget sufficient to proving that an explanation is coherent---subject to a particular use-case.
    \item The research and integration of additional social science sources on theory of XAI communication with humans: Discourse theory~\cite{nla.cat-vn1906176}; collaboration theory~\cite{doi:https://doi.org/10.1002/9781118909997.ch1}; and other cognitive habits in comprehending explanations of behavior, e.g., the least effort principle~\cite{zipf1950-least-effort-principle}, confirmation bias~\cite{nickerson98-confirmation-bias}, and belief bias~\cite{DBLP:conf/acl/GonzalezRS21}.
\end{enumerate*}

\acks{We are grateful to Tim Miller, Been Kim, Sara Hooker, Hendrik Schuff and Sarah Wiegreffe for valuable discussion and feedback, and to Kremena Goranova and her cats Pippa and Peppi for adorable cat and whiskers pictures in \Cref{fig:test,fig:concept-whiskers}. This project has received funding from the European Research Council (ERC) under the European Union’s Horizon 2020 research and innovation programme, grant agreement No. 802774 (iEXTRACT).
}

\appendix
\section{Current XAI Desiderata as Measures of Coherence} \label{app:faithfulness-coherence-alignment}

As mentioned in \Cref{sec:faithfulness}, human society uses coherence as a measure to rely on explanations, due to intractability in proving correctness. Interestingly---though perhaps unsurprisingly---this narrative can also be applied to the development of the relaxed measures of faithfulness in the XAI community. In this appendix section we show that current XAI measures of faithful explanations can be positioned, with some reservation, as measures of coherence.
%
 

\noindentparagraph{\textbf{Neighborhood similarity}\normalfont{~\cite<e.g.,>{melis2018-robustness-of-interpretability,DBLP:journals/corr/abs-2104-08782,DBLP:conf/naacl/DingK21}}}  measures the degree to which similar events are explained similarly. Failure here (i.e., dissimilarity) can be interpreted as a contradiction, under the assumption that the explanation should generalize to examples in the neighborhood. 

This is a relaxed measure of coherence which only tests for contradictions in a neighborhood of contexts, and assumes that the explanation is a proxy for the explainee's mental model.

\noindentparagraph{\textbf{Model similarity}\normalfont{~\cite<e.g.,>{wiegreffe-pinter-2019-attention,DBLP:conf/naacl/DingK21}}}  measures the degree to which two models with similar \textit{behavior} are explained similarly. One can also define measures based on model dissimilarity for models which behave very differently~\cite{DBLP:journals/corr/abs-1810-03292}.

This measure is a variant of the \textit{neighborhood similarity} above, which expands the contradiction search space, and assumes that the two models' explanations will communicate the same mental model to the explainee. 

\noindentparagraph{\textbf{Fidelity}\normalfont{~\cite{DBLP:journals/corr/RibeiroSG16,guidotti-fidelity}}}  measures the degree to which a simpler, ``explainable'' surrogate model is able to mimic the black-box model. In this case, the explanation of the black-box model is the simpler model.\footnote{\textit{Fidelity} can also be considered a special case of model similarity.} 

This measure is a direct adaptation of coherence: The simple model serves as the hypothesis. The budget for proving or refuting coherence can be formalized as the breadth and depth of search for possible instances for which the surrogate model fails to mimic the explained model. However, the required level of fidelity (i.e., quantity of contradictions) is challenging to relate to theory of mind literature. Empirical XAI studies that aim to connect user trust to explanation fidelity found that the way explanations are presented and the underlying model accuracy often overshadow the effect of fidelity, thus making it hard to draw conclusions from the perspective of explainees~\cite{Papenmeier2019HowMA,Larasati2020TheEO}. 

Additionally, some methods of ``surrogate model'' explanations that report fidelity only attempt to mimic the black-box model locally around a particular instance of behavior. Such methods have a weaker connection to coherence, since they do not attempt to fit model behavior across the possible input space. 

\noindentparagraph{\textbf{Relaxed ground truth evaluation}\normalfont{~\cite<e.g.,>{Sippy2020DataSA,DBLP:conf/acl/ZhangHZYCZ20,DBLP:journals/corr/abs-2106-08376,bastings2021will,DBLP:journals/corr/abs-2104-14403}}}  defines a ground truth on ``correct'' explanation by explaining processes which are guaranteed, or are very likely, to reason in a particular way (e.g., a biased model designed to err systematically, or introducing a ``watermark'' to the data which is perfectly correlated with a label; see~\citet{DBLP:journals/corr/abs-2104-14403,bastings2021will}). 

The connection to coherence is straightforward---the explanations are measured via the degree of accuracy to the ground truth---but notably, the empirical budget for proof of coherence manifests in the observed space of AI behavior for which the ground truth exists. For example, evaluating via watermarking only carries real weight for the space of examples with the watermark.

\noindentparagraph{\textbf{Simulatability}\normalfont{~\cite<e.g.,>{DoshiVelez2017TowardsAR,DBLP:conf/emnlp/HaseZXB20,DBLP:journals/corr/abs-2005-01831}}} measures the ability of human explainees to simulate the AI process in a particular setting. 

Simulatability is a sub-case of coherence: Where coherence measures the presence of contradictions to the mental model in all abstract meanings of this definition, simulatability tests for contradictions strictly at the final decision level. Therefore a failure by the user to predict the AI is a clear sign that a contradiction \textit{exists} between the user's mental model and the AI, but it may not be clear what the contradiction is through simulatability alone.

\section{Criticism: On Decision-level (local) and Model-level (global) Explanations} \label{app:criticism}

XAI literature commonly categorizes explanations into two groups: Explaining singular decisions (decision explanations, local explanations) and explaining the entire scope of model behavior (model explanations, global explanations)~\cite{10.1162/tacl_a_00254,DBLP:journals/jair/BurkartH21,DBLP:journals/ai/SetzuGMTPG21}. This gives a taxonomy of explanation mechanisms, unrelated to the mental model of a particular explainee. 


In this appendix, we scrutinize the utility of this categorization: Is the categorization of decision and model explanations potentially descriptive of any differences in the explainee's mental model?

\noindentparagraph{\textbf{Decision-level explanations and coherence.}}
Decision explanations, in themselves, by definition are not constrained with coherence, since they only explain individual instances of behavior. However, this does not mean that they are not \textit{perceived to be} describing generalizing behavior. 

Indeed, under the framework of coherence, explanation is inherently an attempt to communicate generalizing rules. Decision level explanations should be considered as modes of communicating information which can apply beyond the explained instance of behavior.

Given this conclusion, we argue that ``decision-level'' categorization is potentially \textit{misleading} as a description of explanation methods. This argument has also been discussed by~\citet{DBLP:journals/corr/abs-2009-14795}.


\noindentparagraph{\textbf{Is the decision-level and model-level categorization descriptive of the function of XAI methods?}} 
Both decision-level and model-level explanations can communicate information about representation causes, internal representation, external causes, as well as counterfactual and bifactual information directly. However, they aim to explain different events: In decision explanations, the event is the final decision of the AI on a particular instance. 
But model-level explanations can potentially explain two different events:
\begin{enumerate}
    \item The event can be the model itself as the outcome of the process that created it. For example, characterizing the functionality of different components in a compositional neural network~\cite{DBLP:conf/acl/SubramanianBGWS20} or the different kernels in a convolutional neural network~\cite{DBLP:conf/eccv/ZeilerF14} explains the model by building a counterfactual context which would have resulted in a different model. 
    \item The event can be the aggregation of the model's behavior on a large collection of instances, making it an aggregating case of decision-level explanations. For example, 
in explaining that a model achieves strong performance on some task because of exploiting a spurious heuristic~\cite{DBLP:conf/naacl/GururanganSLSBS18}, the ``contrast case'' is a reality where the model is the same, but the \textit{instance space} is different (from instances that exhibit the heuristic, to instances that do not)---such that its \textit{decisions} would be different in this instance space, compared to the previous decisions \cite<e.g.,>{DBLP:journals/corr/abs-2104-08161,DBLP:conf/emnlp/RosenmanJG20,DBLP:conf/acl/McCoyPL19}.
\end{enumerate}

The two different types of events carry different implications on what the explainee may understand about the AI. For example, the contrast case between the two events is different: In (1) it is a different model, while in (2), it is the same model deployed in different contexts.


And yet, the same denomination of ``model-level explanations'' refers to both perspectives interchangeably in the literature~\cite<e.g.,>{DBLP:journals/tetci/ZhangTLT21}. Therefore it can be interpreted as an ambiguous or confusing term, and not descriptive of how the explainee will interpret a given explanation.

\vskip 0.2in
\bibliography{sample}
\bibliographystyle{theapa}

\end{document}